\pdfoutput=1

\documentclass[11pt]{article}

\usepackage[final]{acl}

\usepackage{times}
\usepackage{latexsym}
\usepackage{hyperref}

\usepackage[T1]{fontenc}

\usepackage[utf8]{inputenc}

\usepackage{microtype}

\usepackage{inconsolata}

\usepackage{graphicx}

\usepackage{url}
\usepackage{bbm}
\usepackage{graphicx}
\usepackage{multirow}
\usepackage{amsmath}
\usepackage{amssymb}
\usepackage{mathtools}
\usepackage{amsthm}
\usepackage{microtype}
\usepackage{subfigure}
\usepackage{booktabs} 

\title{CLIP-MoE: Towards Building Mixture of Experts for CLIP with Diversified Multiplet Upcycling}

\author{
 \textbf{Jihai Zhang\textsuperscript{1}},
 \textbf{Xiaoye Qu\textsuperscript{2}},
 \textbf{Tong Zhu\textsuperscript{3}},
 \textbf{Yu Cheng\textsuperscript{1} \thanks{Corresponding author.}},
\\
\\
 \textsuperscript{1}The Chinese University of Hong Kong,
 \textsuperscript{2}Shanghai AI Laboratory,
 \textsuperscript{3}Schoow University
}

\begin{document}
\maketitle
\begin{abstract}
Contrastive Language-Image Pre-training (CLIP) has become a cornerstone in multimodal intelligence. 
However, recent studies discovered that CLIP can only encode one aspect of the feature space, leading to substantial information loss and indistinctive features. 
To mitigate this issue, this paper introduces a novel strategy that fine-tunes a series of complementary CLIP models and transforms them into a \textbf{CLIP-MoE}.
Specifically, we propose a model-agnostic \textbf{Diversified Multiplet Upcycling (DMU)} framework for CLIP. 
Instead of training multiple CLIP models from scratch, DMU leverages a pre-trained CLIP and fine-tunes it into a diverse set with highly cost-effective multistage contrastive learning, thus capturing distinct feature subspaces efficiently.
To fully exploit these fine-tuned models while minimizing computational overhead, we transform them into a CLIP-MoE, which dynamically activates a subset of CLIP experts, achieving an effective balance between model capacity and computational cost. 
Comprehensive experiments demonstrate the superior performance of CLIP-MoE across various zero-shot retrieval, zero-shot image classification tasks, and downstream Multimodal Large Language Model (MLLM) benchmarks when used as a vision encoder.
Codes are released at \href{https://github.com/OpenSparseLLMs/CLIP-MoE}{https://github.com/OpenSparseLLMs/CLIP-MoE}
\end{abstract}

\section{Introduction}
Contrastive Language-Image Pre-training (CLIP) \citep{radford2021learning} 
is a strong vision-language foundation model that utilizes large-scale datasets to learn comprehensive visual representations by bridging vision and language via contrastive image-text pre-training.
{It has been broadly applied in widespread areas such as image} \citep{wang2023exploring,zhang2023biomedclip}, audio~\citep{guzhov2022audioclip}, and video~\citep{rasheed2023fine} understanding, cross-modal retrieval~\citep{ma2022x,zhao2024retrieval}, multimodal generation~\citep{ramesh2022hierarchical,xie2024intelligent}, and data filtering~\citep{schuhmann2022laion}. Recently, CLIP further serves as the vision encoder for various Multimodal Large Language Models (MLLMs)~\citep{alayrac2022flamingo, liu2024llava, liu2024visual, chen2024internvl, li2024mini}.

However, existing CLIP models still exhibit inherent limitations. Recent studies have discovered that CLIP merely encodes a portion of the input's feature space, thus discarding a substantial amount of useful information~\citep{tang2023lemons,tong2024eyes,bleeker2022reducing}.
For instance, when using CLIP as a vision encoder in Multimodal Large Language Models (MLLMs), it frequently produces blind pairs~\citep{tong2024eyes}, where two semantically different images with similar visual components are encoded into the same representation. 
Such indistinctive features severely confuse the reasoning process of MLLM and damage downstream tasks.
To improve the ability of CLIP to capture more distinguished information, remarkable efforts have been made to improve the quality of training data and scale up model size. 
However, these works typically train a new CLIP model from scratch~\citep{li2024if,ma2024mode,xu2023demystifying}, which is resource-intensive.
Meanwhile, an isolated CLIP model may still only encode partial information.  
Therefore, a natural question is raised:
\textit{Can we generate and utilize diverse complementary CLIP models with minimal overhead, without requiring retraining?}

To this end, we propose a \textbf{Diversified Multiplet Upcycling (DMU)} framework for CLIP, 
which constructs a set of complementary CLIP models at a low cost and integrates them using a sparsely activated Mixture of Experts (MoE) architecture. 
MoE has proven effective in scaling model capacity while maintaining fixed activated parameters, enhancing both performance and robustness~\citep{jiang2024mixtral,Dai2024DeepSeekMoETU,chen2024texttt}.
In our proposed DMU framework, instead of training from scratch, we first fine-tune the base CLIP to produce a series of multiplet CLIP models with Multistage Contrastive Learning (MCL)~\citep{zhang2024avoiding}. 
Concretely, MCL encodes diversified information through a multistage clustering and fine-tuning process, generating a CLIP model at each stage and capturing different aspects of the input information. 
Notably, these generated CLIP models share all parameters except for the feed-forward network (FFN) layers during MCL fine-tuning. In this way, we can easily transform them into a \textbf{CLIP-MoE}, which dynamically activates a subset of experts and gets rid of ensembling the CLIP models.  
Finally, through fine-tuning the router in CLIP-MoE, we ensure the full utilization of all experts, enabling CLIP-MoE to capture richer and more distinctive features than the base model, while leveraging sparsity of MoE to avoid the explosion of activated parameters.

We demonstrate that using a small high-quality image-caption dataset, the MCL-initialized CLIP-MoE significantly improves CLIP's performance. 
Notably, on retrieval tasks, CLIP-MoE outperforms the base OpenAI CLIP model by about 20\%, while incurring minimal additional training overhead---less than 2\% of the total computational cost of training the base CLIP model from scratch. When serving as a vision encoder for MLLMs, CLIP-MoE also shows substantial improvements in most benchmarks simply by replacing the original vision encoder. Our experiments show that CLIP-MoE not only outperforms other fine-tuning baselines but also surpasses popular MoE-construction methods such as Sparse Upcycling~\citep{komatsuzaki2022sparse}. 

In summary, the contributions of this work are as follows:
\textit{First}, we introduce a novel Diversified Multiplet Upcycling framework, which generates a set of diversified multiplet CLIP models from an existing dense CLIP model. This approach provides a new and efficient pathway to scale the CLIP foundation model effectively, offering both practical and computational advantages.
\textit{Second}, we demonstrate that our Diversified Multiplet Upcycling framework effectively generates specialized experts, each capturing distinct and diverse useful information. These experts not only encapsulate richer and more nuanced information but also achieve this with significantly reduced computational costs compared to training from scratch.
\textit{Third}, we conduct extensive experiments across a variety of downstream tasks, including retrieval, classification, and serving as a vision encoder for multimodal large language models (MLLMs). Our results show that \textbf{CLIP-MoE} consistently outperforms the original CLIP model and other strong baselines, underscoring its versatility and effectiveness.

\section{Related Works}
\noindent\textbf{Contrastive Learning.}
In contrastive learning, the core objective is to minimize the distance between positives and the anchor while maximizing the distance between negatives and the anchor within the representation space. This objective compels the model to effectively encode sufficient information of the inputs to distinguish anchors from their negatives.
It has become a central technique in self-supervised learning, aiming to learn representations by bringing semantically similar samples closer in the embedding space while pushing dissimilar samples apart~\citep{chen2020simple,he2020momentum}. This approach has been particularly successful in multimodal settings, where models like Contrastive Language-Image Pre-training (CLIP)~\citep{radford2021learning} have emerged as foundational tools. CLIP aligns visual and textual representations by training on vast datasets of paired images and text, enabling the model to bridge different modalities effectively.

Despite its success, CLIP is not without its limitations. 
It lacks the capacity to encode discriminative features adequately, and can only capture a fraction of the information within the feature space~\citep{tang2023lemons,tong2024eyes}. 
To address these limitations, recent works mainly focus on improving the quality of training data~\citep{li2024if,ma2024mode, xu2023demystifying,zhang2024long}. However, most of these approaches require retraining the model from scratch, which is computationally expensive, time-consuming, and not easily extendable when better data becomes available. 
In this paper, we introduce Diversified Multiplet Upcycling (DMU) for CLIP, which transforms a dense CLIP model into a CLIP-MoE through multistage fine-tuning on relatively small datasets. Without retraining, DMU enables capturing diverse and discriminative information while significantly enhancing performance with minimal additional computational overhead.

\noindent\textbf{Mixture-of-Experts.}
The Mixture-of-Experts (MoE) architecture can effectively scale the model capacity with fixed activation parameters ~\citep{Fedus2022MoE-Review}.
For each input token, only top-$k$ best experts are selected to obtain an aggregated representation~\citep{Shazeer2017OutrageouslyLN}.
This sparsity allows MoE models to scale to trillions of parameters while maintaining the computational efficiency~\citep{lepikhin2020gshard,fedus2022switch}.
Benefiting from the large model capacity, the model performance can be improved by large margins~\citep{Rajbhandari2022DeepSpeedMoEAM,Dai2024DeepSeekMoETU}.
Besides, specialized experts in MoE models are good at handling a wide range of tasks~\citep{Shen2023MixtureofExpertsMI,zhu-et-al-2024-ddm,lu2024twin} with high robustness~\citep{chen2024texttt}.

The most important challenge in MoE training is expert construction. Randomly initializing an MoE model and training it from scratch requires substantial resource.
Recently, Sparse Upcycling~\citep{komatsuzaki2022sparse} has been proposed to initialize MoE models by copying Feed-Forward Networks (FFN) from dense models as multiple experts. However, these experts are highly homogeneous, limiting the upper bound of the model's capabilities and leading to suboptimal performance \cite{he2024upcycling}.

In this work, we use multi-stage contrastive learning to initialize the experts for MoE training, which learn distinctive information at each stage. In this way, our MoE model can obtain better optimization and effectively capture complementary features. 

\section{Preliminaries}
\noindent\textbf{Multistage Contrastive Learning (MCL).} MCL~\citep{zhang2024avoiding} is designed to obtain a series of contrastive models, each capturing different and complementary information from the input data through multiple cluster-and-contrastive processes. 
Specifically, at each stage, the learned representations are clustered. In the following stage, for each anchor, negative samples are drawn only from the same accumulated cluster from the previous stages. In this way, the model learns new information beyond what was captured in earlier stages. 
For example, consider a dataset that contains objects with varying shapes, colors, and textures. In the first stage, the contrastive model might focus on learning color information. After clustering, samples within the same cluster will share the same color. In the second stage, since the anchor and its negative samples share the same color, the model is compelled to learn other features, such as texture, to differentiate between them. After clustering in the second stage, samples in the same accumulated cluster will now share both color and texture. Consequently, in the third stage, the model must focus on other attributes, such as shape, to distinguish between samples. After three stages, we obtain three contrastive models, each encoding distinct information: color, texture, and shape.

Formally, let $\boldsymbol{X}=\{\mathbf{x}_i\}_{i=1}^M$ represent a dataset. After training the encoder in the first stage, we obtain encoded representations $\boldsymbol{Z}_0 =\{f_0(\mathbf{x}_i)\}_{i=1}^M$. By clustering $\boldsymbol{Z}_0$, we obtain cluster assignments $\boldsymbol{Y}_0 = \{\mathbf{y}_{(i,0)}\}_{i=1}^M$. In the $j^{th}$ stage, after the cluster-and-contrastive process, each sample $\mathbf{x}_i$ is assigned to an accumulated cluster $\hat{\mathbf{y}}_{(i,j)}=[\mathbf{y}_{(i,0)},\cdots,\mathbf{y}_{(i,j-1)}]$. The objective at the $j^{th}$ stage is:


\begin{align}\label{eq:infornce_multistage}
    \mathcal{L} =
    & \, \mathbb{E}_{\mathbf{x}, \mathbf{x}^+, \{\mathbf{x}_i^- | \hat{\mathbf{y}}_j = \hat{\mathbf{y}}_{(i,j)}^- \}_{i=1}^m} \notag \\
    & \left[
    -\log \frac{e^{s(\mathbf{z}, \mathbf{z}^+)/\tau}}
    {e^{s(\mathbf{z}, \mathbf{z}^+)/\tau}
    + \sum_{i=1}^m e^{s(\mathbf{z}, \mathbf{z}^-_i)/\tau}}
    \right],
\end{align}

where $\hat{\mathbf{y}}_j$ represents the accumulated cluster assignment of the anchor $\mathbf{x}$ at the $j^{th}$ stage; $\hat{\mathbf{y}}_{(i,j)}^-$ 
denotes the accumulated cluster assignment of the negative sample $\mathbf{x}_i^-$ at the $j^{th}$ stage; and \( s(\cdot, \cdot) \) denotes cosine similarity. In our proposed Diversified Multiplet Upcycling, we leverage the MCL framework to fine-tune a base model and extract a series of experts for the MoE, whereas the original MCL results in a series of standalone CLIP models.

\noindent\textbf{Mixture of Experts (MoE).}
Mixture of Experts (MoE) is an efficient architecture designed to scale large models by dynamically routing inputs through a subset of specialized sub-models, or ``experts''. This structure allows the model to maintain high overall capacity while only utilizing a fraction of its parameters for any given input, thereby optimizing both computational efficiency and performance.

In the context of Transformer, an MoE layer~\citep{jiang2024mixtral} typically replaces the standard feed-forward network (FFN) with a set $\{E_i\}_{i=1}^N$ of $N$ experts, each of which is an independent FFN. Given an input token representation $\mathbf{x}$, it 
first passes through a gating network $\mathbf{W}_r$ to obtain the logits corresponding to each expert, then the largest Top-K experts will be chosen, and finally, the probabilities of these selected experts are normalized using Softmax. In this way, we can obtain the probability $R(\mathbf{x})$ of selected experts among all $N$ experts. 


\begin{align}\label{eq:moe}
    \mathbf{x}_{\text{out}} 
    &= \sum_{i=1}^{N} R(\mathbf{x})_i \cdot E_i(\mathbf{x}), \\
    R(\mathbf{x}) 
    &= \text{Softmax}(\text{TopK}(\mathbf{x} \cdot \mathbf{W}_r)).
\end{align}

where $R(\mathbf{x})_i$ denotes the $i$-th routing weight vector produced by the router network $\mathbf{W}_r$. 

To ensure that all experts are utilized effectively and prevent the model from overfitting to a small subset of experts, a load balancing loss~\citep{fedus2022switch} is often added to the primary loss function. This loss penalizes unbalanced expert usage by encouraging a more uniform distribution of input tokens across all experts.

\section{Diversified Multiplet Upcycling for CLIP}
\begin{figure*}[ht]
    \centering
    \includegraphics[width=\textwidth]{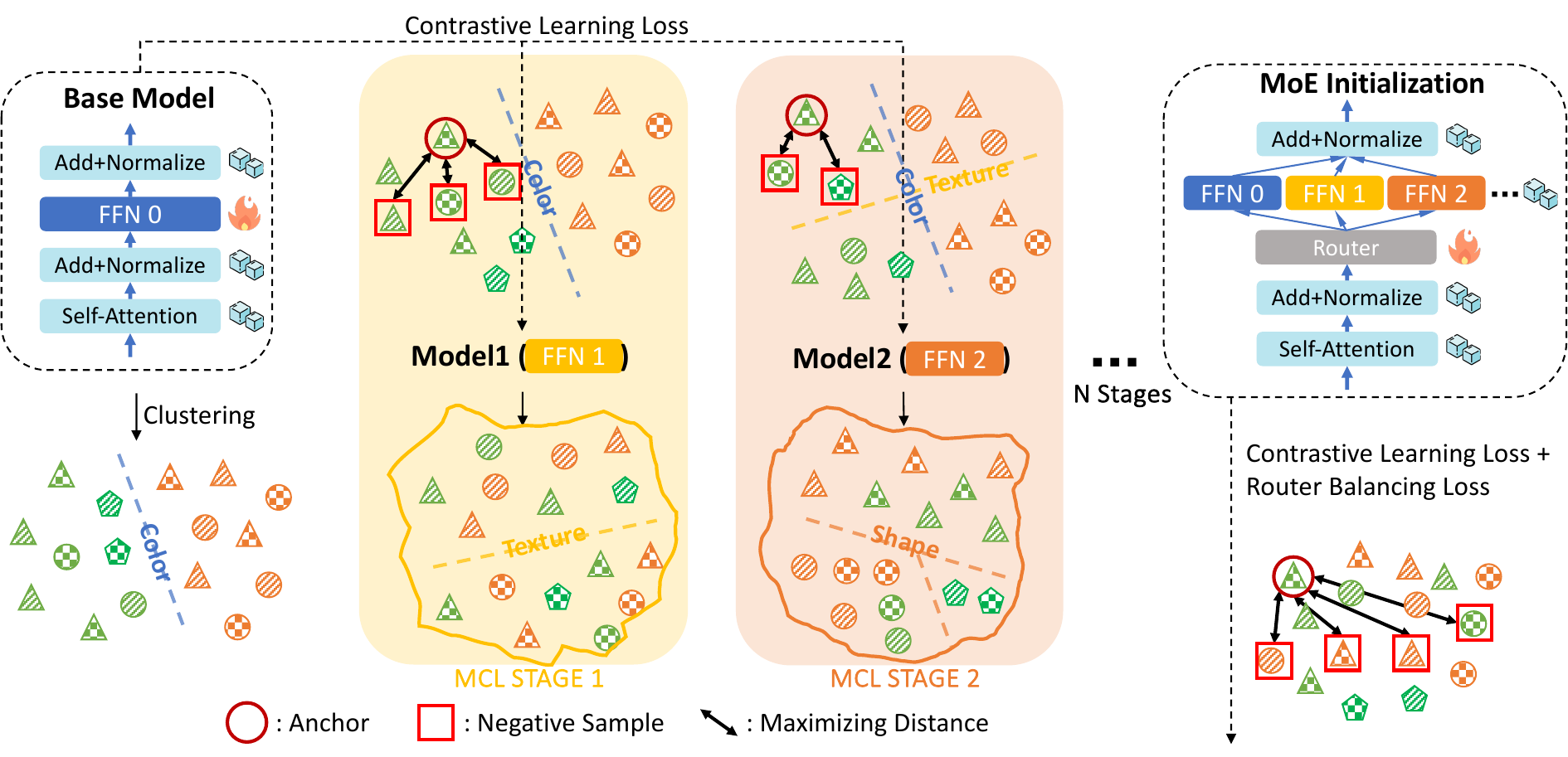}
    \caption{Overview of Diversified Multiplet Upcycling: Our approach involves three key steps. (a) Fine-tuning the base CLIP model using the MCL framework while freezing all parameters except for the FFN layers. This process yields a new set of FFN layers at each stage of MCL. (b) Using the obtained FFN layers as experts to initialize a CLIP-MoE. (c) Continuously fine-tuning the CLIP-MoE using both contrastive learning loss and a router balancing loss to optimize the routers. The terms ‘color’, ‘shape’, and ‘texture’ are metaphorical representations of abstract features.}
    \label{fig:method}
\end{figure*}

\noindent\textbf{Expert Extraction.}
We begin by extracting a series of Feed-Forward Network (FFN) layers utilizing Multistage Contrastive Learning (MCL) to fine-tune a pre-trained base CLIP model for multiple stages. During fine-tuning, we freeze all parameters of the base CLIP model except for the FFN layers within each transformer block in both the image and text encoders. Because the distributions of contrastive negative samples in different MCL stages are distinct, the FFN layers at each stage will learn diversified and complementary information distinct from previous stages. 
For clarity, we use superscripts to index the transformer blocks and subscripts to index the MCL stages or MoE experts. Suppose we are fine-tuning a transformer-based CLIP model, where the image encoder contains $A$ transformer blocks and the text encoder contains $B$ transformer blocks.
The FFN layers in the original base model are denoted as $\{E_0^{(i)}\}_{i=1}^{A+B}$. As illustrated in Figure~\ref{fig:method}, the base model might initially focus on color-related information. During MCL Stage 1, only the FFN layers are fine-tuned. After the cluster-and-contrast process in MCL, the FFN layers $\{E_1^{(i)}\}_{i=1}^{A+B}$ in the fine-tuned model learn new information beyond color, such as texture. In MCL Stage 2, the model further fine-tunes the FFN layers, resulting in $\{E_2^{(i)}\}_{i=1}^{A+B}$, which now encodes additional features such as shape. Through two stages of MCL, we obtain FFN layers where $\{E_0^{(i)}\}_{i=1}^{A+B}$ focus on color, $\{E_1^{(i)}\}_{i=1}^{A+B}$ on texture, and $\{E_2^{(i)}\}_{i=1}^{A+B}$ on shape.

\noindent\textbf{Initialization of Mixture of Experts.}
Once a series of FFN layers $\{E_j^{(i)}\}_{j=0}^N$ have been obtained through $N$ stages of MCL, we utilize these FFNs as the experts in a Mixture of Experts (MoE) model, as depicted in Figure~\ref{fig:method}. According to Equation~\ref{eq:moe}, in the $i^{th}$ transformer block of the base CLIP model, the original FFN layer is replaced with a randomly initialized router and a set of experts:


\begin{align}\label{eq:mcl-moe}
    \mathbf{x}_{\text{out}}^{(i)} 
    &= \sum_{j=0}^{N} R^{(i)}(\mathbf{x}^{(i)})_j \cdot E^{(i)}_j(\mathbf{x}^{(i)}), \\
    R^{(i)}(\mathbf{x}^{(i)}) 
    &= \text{Softmax}(\text{TopK}(\mathbf{x}^{(i)} \cdot \mathbf{W}^{(i)}_r)).
\end{align}

where $R^{(i)}(\mathbf{x})_j$ denotes the $j$-th component of the routing weight vector produced by the router network $\mathbf{W}^{(i)}_r$ in the $i^{th}$ transformer block. This setup results in a CLIP-MoE model where different experts within different transformer blocks specialize in distinct aspects of the input.

\noindent\textbf{Continuous Fine-Tuning of CLIP-MoE.}
To enable the model to learn optimal routing strategies while preserving the information learned by the FFN layers during MCL, we further fine-tune the routers while freezing all other parameters. We apply the standard contrastive learning loss while incorporating an auxiliary load balancing loss, following the approach from \citet{fedus2022switch}, to encourage a balanced load across experts.
Given $N+1$ experts indexed by $j=0$ to $N$, and a batch $\mathcal{B}$ with $T$ tokens, the load balancing loss for the $i^{th}$ transformer block is defined as:

\begin{align}\label{eq:balancing-loss}
    \mathcal{L}_{\text{balance}} &= N \cdot \sum^N_{j=0} f_j \cdot P_j, \\
    f_j &= \frac{1}{T} \sum_{x \in \mathcal{B}} \mathbbm{1}\{\operatorname{argmax} p(x) = j\}, \\
    P_j &= \frac{1}{T} \sum_{x \in \mathcal{B}} p_j(x).
\end{align}
where $f_j$ is the fraction of tokens assigned to expert $j$, and $p(x)$ is the logit output from the router network; $P_j$ represents the fraction of router probability allocated to expert $j$, which is the mean of $p_j(x)$, the probability of routing token $x$ to expert $j$. For simplicity, we omit the transformer block index $i$ in the equation. Since $f_j$ and $P_j$ are positive and both their sums are equal to $1$, $\mathcal{L}_{balancing}$ is minimized if and only if $f_j=\frac{1}{T},\ P_i=\frac{1}{T}$. This balancing loss encourages not only a uniform distribution of actual tokens routed to each expert (i.e., ensuring that all experts have equal importance), but also a uniform distribution of router confidence across tokens (i.e., preventing the router from being overly confident for some tokens and underconfident for others). With this auxiliary load balancing loss, the total loss is given by:
\begin{align}\label{eq:total-loss}
    \mathcal{L}=\mathcal{L}_{CLIP} + \alpha \cdot \frac{1}{A+B}\sum_{i=1}^{A+B}\mathcal{L}^{(i)}_{balance}.
\end{align}
Following \cite{fedus2022switch}, we set $\alpha = 0.01$ by default. By applying MoE-Packing to CLIP, we obtain a CLIP-MoE model that is capable of capturing more useful information than the base model, with minimal computational overhead, resulting in a robust and efficient enhancement of CLIP.

\section{Experiments}
\label{sec:exp}

\subsection{Datasets}
To fully showcase the potential of our MCL-initialized CLIP-MoE, we implement our experiments on the following two image-caption datasets respectively.

\noindent\textbf{Recap-DataComp.}  
Recap-DataComp-1B~\citep{li2024if} is a large-scale dataset comprising 1.3 billion high-quality image-caption pairs. This dataset is derived from the original DataComp-1B dataset, with all images re-captioned using a fine-tuned LLaVA-1.5 model powered by LLaMA-3~\citep{dubey2024llama}. \cite{li2024if} utilized this dataset to train CLIP models from scratch, resulting in significant improvements in retrieval performance. Due to computational constraints, our experiments use a randomly sampled subset of 1 million pairs from Recap-DataComp-1B, referred to as Recap-DataComp-1M, to demonstrate the data efficiency of our proposed pipeline.

\noindent\textbf{ShareGPT4V.}  
ShareGPT4V~\citep{chen2023sharegpt4v} is a high-quality image-text dataset containing 1.2 million highly descriptive captions. The captions are generated by a Multimodal Large Language Model (MLLM) fine-tuned on 100k image-text pairs produced by GPT4V, resulting in well-aligned image-text pairs. 

\subsection{Baselines}
We compare against three approaches: (1) \textbf{Direct fine-tuning} to isolate the performance impact of additional data; (2) \textbf{Sparse Upcycling}~\citep{komatsuzaki2022sparse}, a popular method to efficiently initializes MoE models from dense checkpoints; (3) \textbf{Long-CLIP}~\citep{zhang2024long} that aligns image features with paired short/long captions, though limited to datasets with this specific structure and requiring substantial computation. We also evaluate CLIP-MoE as a vision encoder for \textbf{LLaVA-1.5}~\citep{liu2024improved}, a standard MLLM baseline using a CLIP-to-LLM projection, where we replace its vision encoder with our CLIP-MoE to evaluate representation quality under identical fine-tuning protocols.

\subsection{Training Setup}

By default, we use OpenAI CLIP-ViT-L/14~\citep{radford2021learning} as the base model for our Diversified Multiplet Upcycling approach. During the clustering process at each stage of MCL, we cluster the image features into 3 clusters and the text features into 3 clusters, resulting in 9 clusters per stage (the Cartesian product of the image and text feature clusters). To accommodate longer text inputs, we interpolate the positional embeddings following the approach in \citep{zhang2024long}. The global batch size is maintained at 800 unless otherwise specified. To balance performance and computational cost, we set the number of experts to 4 and use top-2 activation.

\begin{table}[!ht]
\caption{Performance of different experts across various attributes in MMVP. The highest value for each attribute is highlighted.}
\centering
\footnotesize
\setlength{\tabcolsep}{5.8pt}
\begin{tabular}{lcccc}
\toprule
\textbf{Attribute} & \textbf{Expert0} & \textbf{Expert1} & \textbf{Expert2} & \textbf{Expert3} \\
\midrule
O\&D & 40 & 33.3 & \textbf{46.7} & \textbf{46.7} \\
PSF & \textbf{33.3} & 26.7 & 26.7 & 13.3 \\
S\&C & 20 & 40 & \textbf{53.3} & 40 \\
Q\&C & \textbf{60} & 46.7 & 40 & 40 \\
P\&R & \textbf{46.7} & 33.3 & 40 & 26.7 \\
C\&A & \textbf{26.7} & 13.3 & 6.7 & 6.7 \\
S\&P & 26.7 & \textbf{46.7} & 40 & 33.3 \\
Texts & 26.7 & 40 & \textbf{46.7} & 40 \\
V\&P & 53.3 & 46.7 & 40 & \textbf{60} \\
\bottomrule
\end{tabular}
\label{tab:mmvp}
\end{table}

\begin{table*}[!htbp]
\caption{Performance comparison on image-to-text (I2T) and text-to-image (T2I) retrieval tasks using the COCO and Flickr30k datasets. The models were trained and evaluated on the Recap-DataComp-1M (Recap-DC) and ShareGPT4V (ShareGPT) datasets, respectively. The best performance for each dataset is highlighted in bold. Our CLIP-MoE consistently outperforms all baselines across all tasks.}
\centering
\footnotesize
\begin{tabular}{l|l|ccc|ccc|ccc|ccc}
\toprule
 & &   \multicolumn{3}{c|}{\textbf{COCO I2T}} & \multicolumn{3}{c|}{\textbf{COCO T2I}} & \multicolumn{3}{c|}{\textbf{Flickr I2T}} & \multicolumn{3}{c}{\textbf{Flickr T2I}} \\
\textbf{Dataset}& \textbf{Model}  & \textbf{@1} & \textbf{@5} & \textbf{@10} & \textbf{@1} & \textbf{@5} & \textbf{@10} & \textbf{@1} & \textbf{@5} & \textbf{@10} & \textbf{@1} & \textbf{@5} & \textbf{@10} \\
\midrule
& OpenAI &  56.1 & 79.5 & 86.8 & 35.4 & 60.1 & 70.2 & 48.5 & 72.6 & 80.8 & 28.0 & 49.3 & 58.7 \\
\midrule
\multirow{3}{*}{\textbf{Recap-DC}} 
& Direct FT & 58.9  & 81.5 & 88.5 & 44.3 & 69.5 &78.8  & 41.6 & 66.5 & 76.1 &37.2  & 60.4 & 69.5 \\
&Upcycling  & 59.2  & 81.7 & 88.7 & \textbf{45.8} & \textbf{70.9} & \textbf{79.9} & 42.1 & 67.3 & 77.0 & 39.4 & 62.9 &71.7  \\
&CLIP-MoE & \textbf{64.0}  & \textbf{85.1} &\textbf{ 90.8} & 45.2 & 70.2 & 79.4 & \textbf{56.8} &\textbf{ 80.1} &\textbf{87.0}  &\textbf{40.8}  & \textbf{63.8} & \textbf{72.5} \\
\midrule
\multirow{3}{*}{\textbf{ShareGPT}} 
&Direct FT & 63.3  & 84.9 & 91.0 & 44.5 & 70.0 & 78.9 &50.5  &74.4  &82.3  &38.5  & 61.3 &69.9  \\
& Upcycling& 62.9  & 84.6 & 90.8 &45.2  & 70.6 &79.6  & 49.6 & 73.8 &82.1  &39.5  & 62.4& 71.1 \\
&Long-CLIP & 62.8 & 85.1 & 91.2 & 46.3 & 70.8 & 79.8 & 53.4 & 77.5 & 85.3 & 41.2 & 64.1 & 72.6\\
&CLIP-MoE &  \textbf{65.0} & \textbf{86.0} & \textbf{92.0} & \textbf{46.8} & \textbf{71.7} & \textbf{80.4} & \textbf{60.5} & \textbf{82.3} & \textbf{88.8} & \textbf{42.1} & \textbf{64.7} & \textbf{73.2} \\
\bottomrule
\end{tabular}
\label{tab:retrieval}
\end{table*}

\subsection{Training Cost}
We use 8 A100 GPUs for training. To train the CLIP-MoE model with four experts, we introduce three additional MCL fine-tuning stages, each trained for 1 epoch. When using the ShareGPT4V dataset, each MCL stage takes approximately 0.5 hours, and the router fine-tuning stage also takes about 0.5 hours. In total, the training time is less than 2.5 hours. In comparison, Long-CLIP training under the same conditions takes around 6 hours, making our approach significantly more efficient. Our maximum GPU memory usage is 8×65955MB, which is comparable to Long-CLIP's 8×63581MB. When training on the Recap-DataComp-1M dataset, the training cost is even lower. During inference, with top-2 activation, the activated parameter size of our CLIP-MoE is approximately 1.7 times that of the base model (OpenAI CLIP-ViT-L/14).

\begin{table*}[!ht]
\caption{Performance comparison between OpenAI CLIP and CLIP-MoE as vision encoders in LLaVA1.5. The best performance for each dataset is highlighted in bold.}
\centering
\footnotesize
\setlength{\tabcolsep}{4.2pt}
\begin{tabular}{lcccccccccc}
\toprule
\textbf{Method} & \textbf{MME} & \textbf{POPE} & \textbf{MMBench} & \textbf{MM-Vet} & \textbf{VisWis} & \textbf{MMStar} & \textbf{OCRBench} & \textbf{VQAv2} & \textbf{TextVQA} & \textbf{GQA}  \\
\midrule
OpenAI CLIP & \textbf{1510.7} & 85.9 & 64.3 & 30.6 & 54.4 & 33.3 & 31.2 & 78.5 & 46.1 & 62.0 \\
CLIP-MoE & 1486.2 & \textbf{86.4} & \textbf{66.1} & \textbf{31.5} & \textbf{56.5} & \textbf{34.1} & \textbf{31.8} & \textbf{79.2} & \textbf{46.8} & \textbf{62.6} \\
\midrule
OpenAI CLIP & 1522.6 & 85.9 & 67.7 & 35.3 & 56.7 & 36.1 & 33.6 & {80.0} & \textbf{48.7} & 63.2 \\
CLIP-MoE & \textbf{1560.1} & \textbf{86.5} & \textbf{69.3} & \textbf{39.5} & \textbf{59.2} & \textbf{36.7} & \textbf{34.4} & \textbf{80.0} & 48.3 & \textbf{63.8} \\
\bottomrule
\end{tabular}
\label{tab:mllm}
\end{table*}


\subsection{Evaluation}
We begin by evaluating whether different experts do capture different usefult information as we expected. Then we evaluate the performance of CLIP-MoE on Zero-Shot Image-Text Retrieval, a key task for assessing whether the CLIP model can capture rich fine-grained information, following \cite{zhang2024long}. All baselines are trained and compared using the Recap-DataComp-1M (Recap-DC) and ShareGPT4V (ShareGPT) datasets, with the exception of Long-CLIP. Long-CLIP is incompatible with the Recap-DataComp dataset, as it requires both a short and long caption for each image, whereas Recap-DataComp provides only one caption per image.
Next, we assess the effectiveness of CLIP-MoE as a vision encoder within LLaVA-1.5, a representative Multimodal Large Language Model (MLLM). LLaVA-1.5 serves as an effective visual representation evaluator, helping to mitigate potential biases present in traditional evaluation tasks~\citep{tong2024cambrian}. Finally, we test CLIP-MoE on traditional Zero-Shot Image Classification tasks, which rely more on coarse-grained features.

\noindent\textbf{Specialization of Experts.}
To investigate whether different experts learn distinct features, we evaluate each expert's performance individually on the MMVP Benchmark~\cite{tong2024eyes}. MMVP requires the CLIP model to select the correct image based on a textual statement from a pair of visually similar images. The evaluation data are carefully filtered into nine distinct attributes by human annotators.
The results in Table~\ref{tab:mmvp} clearly show that different experts specialize in different attributes. For example, Expert0 performs best on attributes such as Presence of Specific Features, Quantity and Count, Color and Appearance, and Viewpoint and Perspective. Expert1 excels in Structural and Physical Characteristics. Expert2 focuses on Orientation and Direction, State and Condition, and Texts, while Expert3 specializes in Orientation and Direction, as well as Viewpoint and Perspective.
These results highlight the effectiveness of our proposed Diversified Multiplet Upcycling, as it successfully generates experts that specialize in capturing diverse and complementary information.

\noindent\textbf{Zero-Shot Image-Text Retrieval.} 
Following the methodology outlined in \cite{zhang2024long}, we evaluate text-to-image (T2I) and image-to-text (I2T) retrieval on the 5k COCO validation set~\citep{lin2014microsoft} and the 30k Flickr30k~\citep{young2014image} dataset. The results are presented in Table~\ref{tab:retrieval}. Given that both Recap-DataComp-1M and ShareGPT4V datasets offer higher caption quality and longer average caption lengths compared to web datasets, Direct Fine-Tuning, Sparse Upcycling, and CLIP-MoE demonstrate superior performance over the original OpenAI model across most tasks, including COCO I2T, COCO T2I, and Flickr T2I. However, for Flickr I2T, Sparse Upcycling, and Direct Fine-Tuning show significant performance degradation on the Recap-DC dataset. In this fine-tuning context, Sparse Upcycling only provides a limited advantage over Direct Fine-Tuning. Although Long-CLIP clearly outperforms both Direct Fine-Tuning and Sparse Upcycling, it is incompatible with the Recap-DataComp dataset, because it requires each image to have both a short and a long caption. In contrast, our proposed CLIP-MoE surpasses all baselines on most tasks across two datasets, maintaining consistent performance by leveraging the diverse information extracted by MoE experts.

\begin{table*}[!ht]
\caption{Ablation study on the impact of MCL expert extraction in CLIP-MoE performance.}
\setlength{\tabcolsep}{4pt}
\centering
\begin{tabular}{lccccccccccccc}
\toprule
 & \multicolumn{1}{c}{\textbf{ImageNet}} & \multicolumn{3}{c}{\textbf{COCO I2T}} & \multicolumn{3}{c}{\textbf{COCO T2I}} & \multicolumn{3}{c}{\textbf{Flickr I2T}} & \multicolumn{3}{c}{\textbf{Flickr T2I}} \\
\cmidrule(lr){2-2} \cmidrule(lr){3-5} \cmidrule(lr){6-8} \cmidrule(lr){9-11} \cmidrule(lr){12-14}
\textbf{Method} & \textbf{Top-1} & \textbf{@1} & \textbf{@5} & \textbf{@10} & \textbf{@1} & \textbf{@5} & \textbf{@10} & \textbf{@1} & \textbf{@5} & \textbf{@10} & \textbf{@1} & \textbf{@5} & \textbf{@10} \\
\midrule
w/o MCL & 75.4 & 62.6 & 84.2 & 90.3 & 43.4 & 68.3 & 77.8 & 56.4 & 79.3 & 86.3 & 37.6 & 60.3 & 69.3 \\
CLIP-MoE & 74.6 & 65.0 & 86.0 & 92.0 & 46.8 & 71.7 & 80.4 & 60.5 & 82.3 & 88.8 & 42.1 & 64.7 & 73.2 \\
\bottomrule
\end{tabular}
\label{tab:ablation}
\end{table*}

\noindent\textbf{Performance in LLaVA-1.5.}
We further evaluate CLIP-MoE as the vision encoder within the LLaVA-1.5 model. The original vision encoder for LLaVA-1.5 is OpenAI's CLIP-ViT-L/14@336px~\citep{radford2021learning}, which is trained on images with a resolution of 336x336 pixels.
To ensure a fair comparison, we use OpenAI's CLIP-ViT-L/14@336px as the base model for MCL and train our CLIP-MoE on the ShareGPT4V dataset at the same 336x336 resolution. After obtaining CLIP-MoE, we freeze it as the vision encoder and follow the same two-stage training procedure as LLaVA-1.5, using Vicuna~\citep{vicuna2023} as the base LLM. We evaluate the MLLMs on ten popular independent MLLM benchmarks~\citep{hudson2019gqa,liu2025mmbench,fu2023mme,chen2024we,yu2023mm,liu2024ocrbench,li2023evaluating,gurari2018vizwiz,singh2019towards,goyal2017making}. As shown in Table~\ref{tab:mllm}, simply replacing the vision encoder with CLIP-MoE yields notable performance improvements across most downstream tasks, with particularly strong gains on MMBench (+1.6), MM-Vet (+4.2), and VizWiz (+2.5). Interestingly, the 13B model even exhibits a larger performance boost than the 7B model, suggesting that larger base LLMs can better leverage the discriminative information captured by CLIP-MoE.
These results strongly support the conclusion that CLIP-MoE extracts richer, more distinctive information from image inputs and encodes higher-quality visual representations, ultimately enhancing the performance of MLLMs.

\begin{table}[!ht]
\setlength{\tabcolsep}{2pt}
\caption{Performance comparison on zero-shot image classification. The models were trained and evaluated on the Recap-DC and ShareGPT4V datasets, respectively. The best performance for each dataset is highlighted in bold.}
\centering
\scriptsize 
\begin{tabular}{l|l|c|c|c|c|c}
\toprule
\textbf{Dataset}& \textbf{Model} & ImgNet & ImgNetO & ImgNetV2 & Cifar10 & Cifar100   \\
\midrule
& OpenAI &  \textbf{75.5} & 31.9 & \textbf{69.9} & 95.4 & 76.8 \\
\midrule
\multirow{3}{*}{\textbf{Recap-DC}} 
& Direct FT & 57.0  & \textbf{32.8} & 51.3 & 91.6 & 68.7 \\
&Upcycling &  61.1 & 32.3 & 55.3 &  93.6&  71.0 \\
&CLIP-MoE & 74.3  & 32.2 & 68.7 & \textbf{95.5} &  \textbf{79.3} \\
\midrule
\multirow{3}{*}{\textbf{ShareGPT}} 
&Direct FT & 59.8  & \textbf{34.5} & 53.3 & 87.8 & 63.1 \\
&Upcycling &  62.5 & 34.4 &56.5  & 91.3 &  67.5 \\
&Long-CLIP & 73.5  & 33.7 & 67.9 & 95.3 &78.5\\
&CLIP-MoE & 74.6  & 33.5 & 68.5 & \textbf{95.7} &\textbf{79.6}\\
\bottomrule
\end{tabular}
\label{tab:classification}
\end{table}

\noindent\textbf{Zero-Shot Image Classification.} 
For a more comprehensive study, we evaluate our CLIP-MoE on the zero-shot image classification accuracy on ImageNet~\citep{deng2009imagenet}, ImageNet-O~\citep{hendrycks2021natural}, ImageNet-V2~\citep{recht2019imagenet}, CIFAR-10~\citep{krizhevsky2009learning}, and CIFAR-100~\citep{krizhevsky2009learning}. The results, presented in Table~\ref{tab:classification}, reveal that no model significantly surpasses OpenAI CLIP in classification accuracy.
We attribute this to two key reasons. First, data limitations: both the Recap-DataComp and ShareGPT4V datasets contain roughly 1M samples, significantly smaller than the 400M samples used to train OpenAI CLIP. This scale difference contributes to overfitting and limited generalization. Second, the nature of classification tasks: coarse-grained features play a dominant role in classification, whereas the fine-grained information captured by the model does not always translate to improved classification accuracy and, in some cases, may even degrade performance.
For instance, Long-CLIP, which learns more fine-grained representations from enhanced and lengthier image captions, improves retrieval performance but exhibits a performance drop on ImageNet and ImageNet-V2. However, CLIP-MoE mitigates this degradation more effectively than Long-CLIP, which explicitly incorporates short captions to preserve coarse-grained feature encoding. Moreover, CLIP-MoE even surpasses OpenAI CLIP on ImageNet-O and CIFAR, suggesting that our proposed DMU approach not only enhances the model’s ability to capture fine-grained information but also maintains coarse-grained feature extraction, ultimately improving overall representation quality.




\noindent\textbf{Ablation Study on MCL Expert Extraction.}
To further evaluate the effectiveness of expert extraction via MCL in Diversified Multiplet Upcycling, we conducted an ablation study on the ShareGPT4V dataset. Specifically, we integrated the original OpenAI CLIP and a CLIP model with FFN layers directly fine-tuned on ShareGPT4V into a vanilla MoE model with two experts.
As shown in Table~\ref{tab:ablation}, CLIP-MoE consistently outperforms the vanilla MoE model (without MCL expert extraction) on retrieval tasks. This highlights the effectiveness of MCL stages in producing experts that capture more meaningful and diverse information. The slight decrease in ImageNet zero-shot classification performance is expected, as not all additional information learned through MCL benefits classification tasks, which tend to depend more on coarse-grained features~\citep{zhang2024long}.

\section{Conclusion}


In this paper, We propose a novel Diversified Multiplet Upcycling framework to construct CLIP-MoE, leveraging multi-stage contrastive learning to extract diverse, complementary experts with minimal computation overhead. 
Instead of ensembling, these experts are integrated through an MoE architecture, capturing richer and more distinctive information from the inputs, while maintaining fixed activation parameters. 
By fine-tuning an off-the-shelf CLIP with a small, high-quality dataset, our method enhances performance without the cost of training from scratch. Our approach is easy to apply, model-agnostic, and provides a new path to scale and improve CLIP foundation models. 


\section*{Limitations}

First, the current experiments are constrained to image and text modalities. While these modalities provide a strong foundation, we aim to expand our method to encompass additional modalities, such as audio and video, to explore its versatility in multimodal learning scenarios.
Second, our evaluation is currently limited to fine-tuning settings. To better understand the scalability and robustness of Diversified Multiplet Upcycling, we plan to experiment with larger datasets and investigate large-scale continuous training regimes. Such experiments will help us further delineate the performance boundaries and practical applicability of our approach.
Finally, although we have successfully tested CLIP-MoE as a vision encoder for multimodal language models (MLLMs), its potential as a text encoder in generative tasks remains underexplored. For instance, integrating CLIP-MoE into frameworks like stable diffusion could open new avenues for improving text-driven generation tasks. 

\bibliography{custom}

\appendix

\section{Appendix}

\begin{figure*}[!ht]
    \centering
    \includegraphics[width=0.99\textwidth]{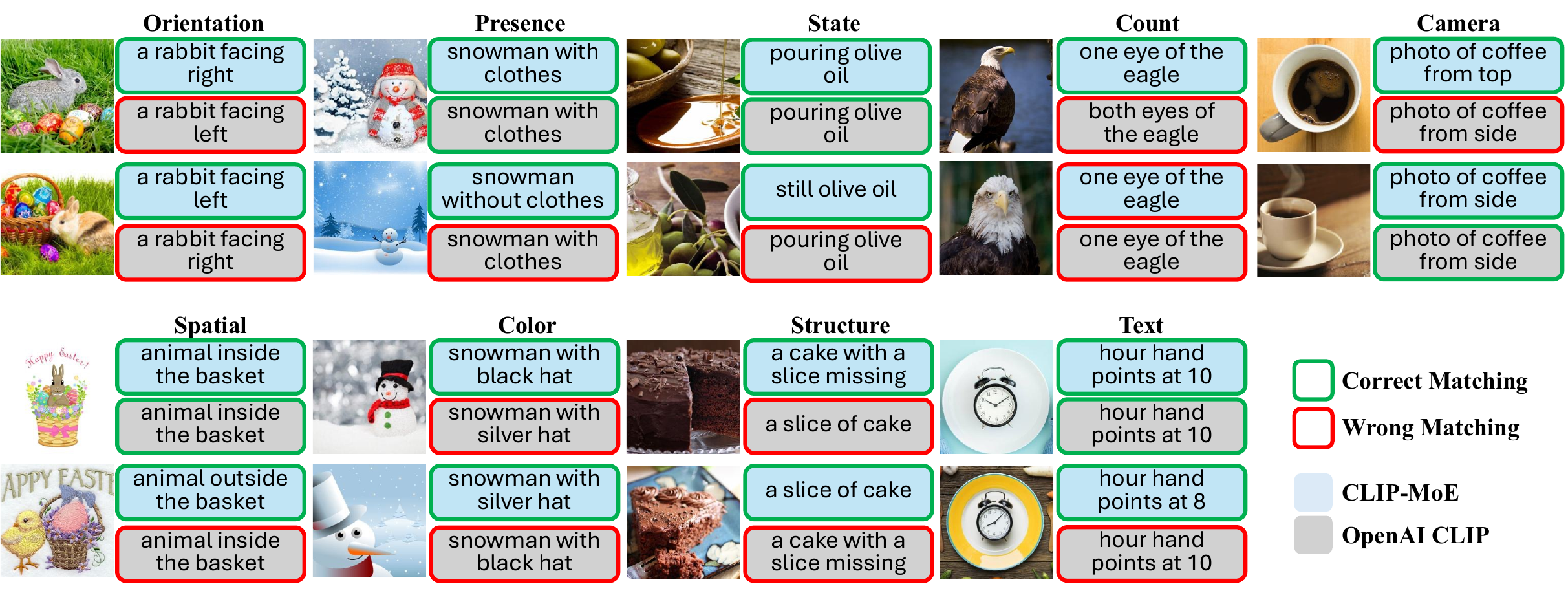}
    \caption{Example cases comparing the performance of CLIP-MoE and OpenAI CLIP on the MMVP-VLM Benchmark, illustrating differences in their ability to capture fine-grained semantic information.}
    \label{fig:case}
\end{figure*}

\begin{figure*}[!ht]
    \centering
    \includegraphics[width=\textwidth]{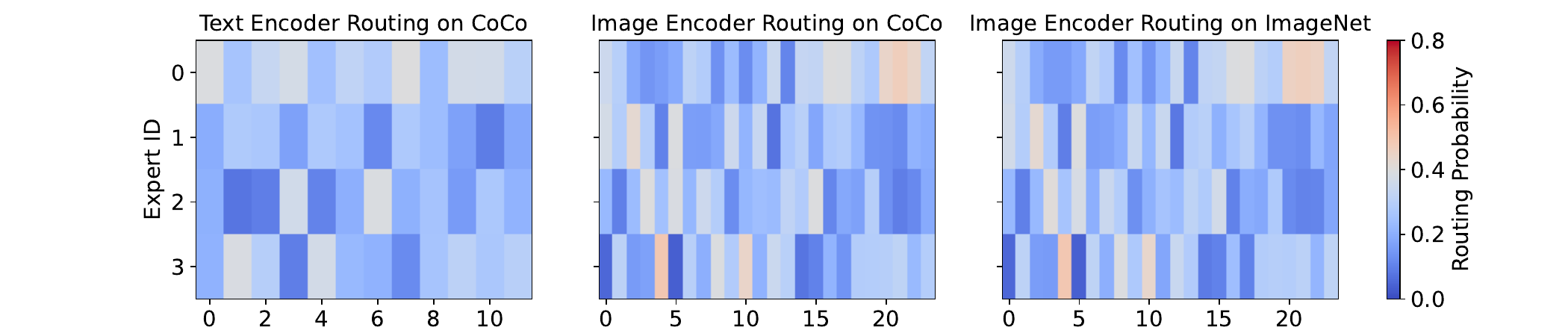}
    \caption{Proportion of tokens assigned to each expert on the COCO and ImageNet validation dataset. Here, we consider experts that are either selected as a first or second choice by the router.}
    \label{fig:routing}
\end{figure*}

\subsection{Case Study.}
We demonstrate the comparison between CLIP-MoE and OpenAI CLIP on samples from the MMVP-VLM Benchmark~\citep{tong2024eyes}. MMVP-VLM contains manually filtered image pairs with different semantics that are difficult to distinguish using the vanilla OpenAI CLIP. We task the models with matching the corresponding statement to the image. As shown in Figure~\ref{fig:case}, OpenAI CLIP struggles to distinguish fine-grained details in these image pairs. In cases like the alarm clock, OpenAI CLIP matches both images to the statement ``hour hand points at 10.'' In other cases, such as the rabbit pair, OpenAI CLIP completely misinterprets the information and matches the opposite statement. However, CLIP-MoE captures more fine-grained details and makes the correct match in most cases. It can accurately capture camera perspectives, as seen in the coffee example, orientation information in the rabbit example, and it demonstrates a superior ability to distinguish relations between objects, such as differentiating between ``animal inside the basket'' and ``animal outside the basket.''

\subsection{Computation and Data Efficiency.}
We compare the performance gains of our CLIP-MoE, trained on a 1M randomly sampled subset of Recap-DataComp-1B, to the CLIP-ViT-L-16-HTxt-Recap~\citep{li2024if}, which was trained from scratch on the entire Recap-DataComp-1B dataset. The activated parameter size of our CLIP-MoE, with 4 experts and top-2 routing, is 0.69B, which is comparable to the 0.64B parameter size of CLIP-ViT-L-16-HTxt-Recap. Thanks to MoE-Packing and leveraging the OpenAI CLIP dense checkpoint, our total training computation cost is less than $2\%$ of that for CLIP-ViT-L-16-HTxt-Recap. As shown in Table~\ref{tab:gain-retrieval}, CLIP-MoE demonstrates comparable performance gains on retrieval tasks relative to CLIP-Recap, with even superior text-to-image retrieval performance on the Flickr30k dataset, highlighting the efficiency of our proposed Diversified Multiplet Upcycling for CLIP. It is worth noting that CLIP-Recap uses an even larger text encoder.

\begin{table}[!ht]
\setlength{\tabcolsep}{3pt}
\caption{Performance gain of CLIP-MoE and CLIP-Recap compared to the OpenAI CLIP-ViT-L-14 on retrieval tasks. }
\centering
\scriptsize
\begin{tabular}{l|cc|cc|cc|cc}
\toprule
 &    \multicolumn{2}{c|}{\textbf{COCO I2T}} & \multicolumn{2}{c|}{\textbf{COCO T2I}} & \multicolumn{2}{c|}{\textbf{Flickr I2T}} & \multicolumn{2}{c}{\textbf{Flickr T2I}} \\
 \textbf{Model}  & \textbf{@1} & \textbf{@5} & \textbf{@1} & \textbf{@5}  & \textbf{@1} & \textbf{@5}  & \textbf{@1} & \textbf{@5} \\
\midrule
CLIP-MoE &  +7.9	&+5.6	&	+9.8&	+10.1&		+8.3&	+7.5&		+12.8&	+14.5	 \\
CLIP-Recap &+10.8&	+7.7&		+12.3	&+12.3&		+10.9&	+8.3&	+11.9	&+12.9\\
\bottomrule
\end{tabular}
\label{tab:gain-retrieval}
\end{table}

\subsection{Routing analysis}
To evaluate whether all the experts learned through MCL are utilized by CLIP-MoE, we perform an analysis of the routing strategy. We use the CLIP-MoE model with 4 experts and top-2 routing trained on ShareGPT4V, and compute the proportion of tokens assigned to each expert. For retrieval tasks, we use the COCO validation dataset, and for zero-shot image classification, we use the ImageNet validation dataset. The analysis results are presented in Figure~\ref{fig:routing}. From the results, we observe that for experts from each MCL stage (represented by each column in the heatmap), there are consistently yellow areas (indicating heavily utilized experts). No column is entirely dark blue, which indicates that all MCL stages contribute useful experts to CLIP-MoE. This further validates the effectiveness of our Diversified Multiplet Upcycling.

\subsection{Artifact Documentation}
We used the pre-trained CLIP model~\citep{radford2021learning} strictly for research purposes, adhering to its original license restrictions.
The primary scientific artifact of this work is the \textbf{Diversified Multiplet Upcycling framework}, a novel methodological contribution for scaling CLIP-based models. While this work does not release new datasets or pre-trained models, the framework itself constitutes a reusable and well-documented artifact designed for cross-modal learning tasks. The framework is applicable to domains such as image-text retrieval, classification, and vision encoding for multimodal large language models (MLLMs), inheriting the language support of the original CLIP model (e.g., English) and extending compatibility to text inputs in multiple languages if the base CLIP supports them. It is validated on tasks including zero-shot classification, image-text retrieval, and MLLM vision encoding (e.g., for stable diffusion). The framework is designed for research purposes only and must adhere to the licensing terms of the original CLIP model, with derivative works (e.g., fine-tuned CLIP-MoE models) required to comply with the same restrictions. Key hyperparameters, such as contrastive learning stages and MoE routing strategies, are described in Section~\ref{sec:exp}, and the modular design ensures reproducibility by following the architectural and training guidelines provided.

\end{document}